\documentclass[11pt]{article}

\usepackage[final]{acl}

\usepackage{times}
\usepackage{latexsym}

\usepackage[T1]{fontenc}

\usepackage[utf8]{inputenc}
\usepackage{microtype}
\usepackage{inconsolata}
\usepackage{graphicx}
\usepackage{url}

\usepackage{amsmath}      
\usepackage{amssymb}      
\usepackage{bm}           
\usepackage{mathtools}    
\usepackage{enumitem}     

\usepackage{amsthm}
\theoremstyle{plain}

\theoremstyle{definition} 

\theoremstyle{remark} 

\usepackage[most]{tcolorbox}
\tcbuselibrary{breakable, skins, theorems, listings} 
\usepackage[dvipsnames]{xcolor} 
\usepackage{cleveref}     
\usepackage{booktabs}     
\usepackage{xcolor}       
\usepackage{multirow}
\usepackage{subcaption}
\usepackage{kotex}
\usepackage{graphicx}
\usepackage{wrapfig}

\title{How Training Data Shapes the Use of Parametric and In-Context Knowledge in Language Models}

\author{
  \textbf{Minsung Kim}$^{1}$,
  \textbf{Dong-Kyum Kim}$^{2}$,
  \textbf{Jea Kwon}$^{2}$,
  \textbf{Nakyeong Yang}$^{1}$ \\
  \textbf{Kyomin Jung}$^{1,\dagger}$,
  \textbf{Meeyoung Cha}$^{2,\dagger}$ \\
  $^{1}$Seoul National University,
  $^{2}$Max Planck Institute for Security and Privacy
  \\
  \texttt{\{kms0805, kjung\}@snu.ac.kr} \quad  \texttt{mia.cha@mpi-sp.org}
}

\begin{document}
\maketitle
\renewcommand{\thefootnote}{\fnsymbol{footnote}}
\footnotetext[2]{Corresponding authors.}
\renewcommand{\thefootnote}{\arabic{footnote}}

\begin{abstract}
Large language models leverage both parametric knowledge acquired during pretraining and in-context knowledge provided at inference time. Crucially, when these sources conflict, models arbitrate based on their internal confidence, preferring parametric knowledge for high-confidence facts while deferring to context for less familiar ones. However, the training conditions that give rise to these fundamental behaviors remain unclear. Here we conduct controlled experiments using synthetic corpora to identify the specific data properties that shape knowledge utilization. Our results reveal a counterintuitive finding: the robust, balanced use of both knowledge sources is an emergent property that requires the co-occurrence of three factors typically considered detrimental, including (i) intra-document repetition, (ii) a moderate degree of intra-document inconsistency, and (iii) a skewed knowledge distribution. We further show that these dynamics arise in real-world language model pretraining and analyze how post-training procedures reshape arbitration strategies. Together, our findings provide empirical guidance for designing training data that supports the reliable integration of parametric and in-context knowledge in language models.\footnote{Code available at \url{https://github.com/kms0805/how-training-data-shapes-pk-ick}.}
\end{abstract}

\section{Introduction}

Large language models (LLMs)~\citep{touvron2023llama, brown2020language, biderman2023pythia} encode vast amounts of world knowledge within their parameters during pretraining~\citep{roberts2020much, petroni2019language, geva2020transformer}. However, reliance on this parametric knowledge is fundamentally limited: it becomes outdated as the world changes and lacks coverage of rare or domain-specific information. To address these limitations, retrieval-augmented generation (RAG)~\citep{lewis2021retrieval, ram2023incontext, shi2023replug} provides external documents as in-context knowledge at inference time. Through this paradigm, models can effectively utilize information that lies outside their parameters, such as contemporary facts or domain-specific details. 

LLMs acquire the ability to leverage both knowledge sources through standard next-token prediction on web corpora~\citep{radford2018improving}, without requiring explicit fine-tuning for retrieval-augmented generation~\citep{ram-etal-2023-context, mallen-etal-2023-trust, shi2023replug}. When these two sources conflict, models do not blindly follow in-context knowledge, which may itself be noisy or incorrectly retrieved. Instead, they exhibit confidence-dependent arbitration: they prefer their internal parametric knowledge for high-confidence facts (i.e., high-probability, low-entropy predictions) while deferring to in-context knowledge for less familiar information~\citep{wu2024clasheval, yu-etal-2023-characterizing}.
Despite the widespread deployment of these systems, the specific training data properties that give rise to these fundamental behaviors remain poorly understood.

\begin{figure*}[t]
\vspace{-8pt}
\centering
\includegraphics[width=\textwidth]{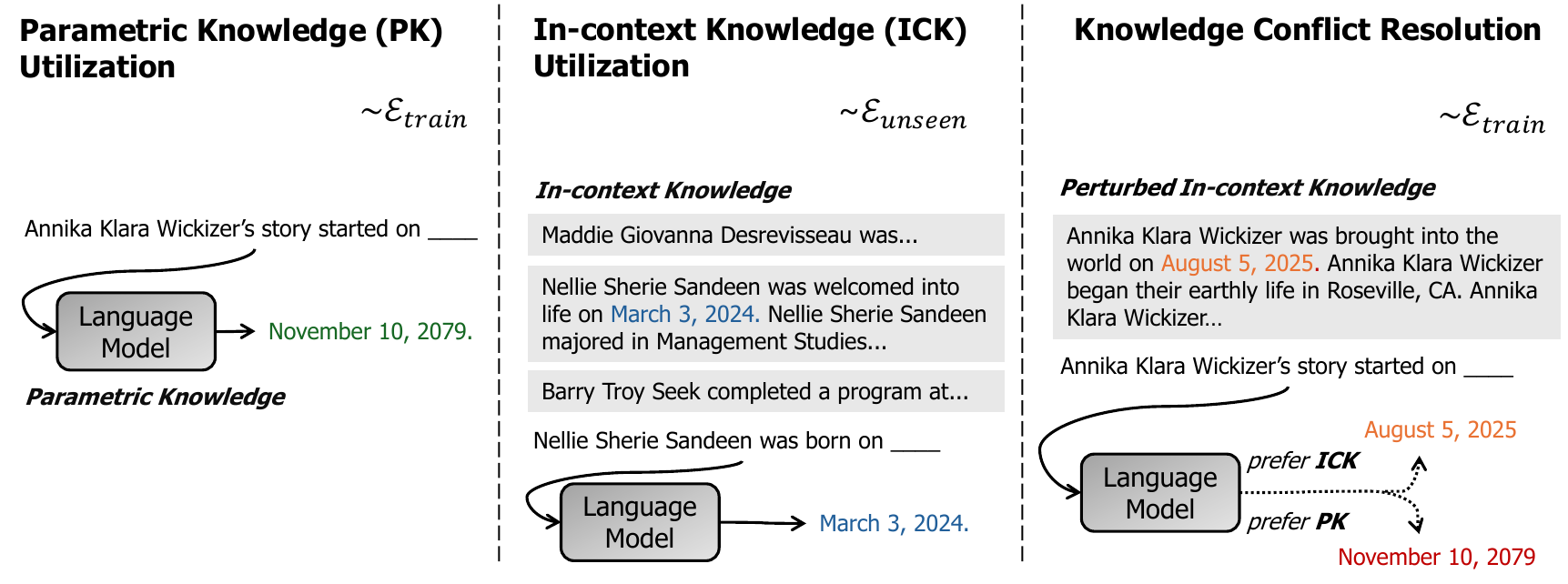}
\caption{Evaluation framework for knowledge utilization and conflict resolution. \textbf{(Left)} Parametric knowledge utilization, where the model recalls knowledge encoded in its parameters about entities seen during training. \textbf{(Middle)} In-context knowledge utilization, where the model extracts and uses knowledge provided only in the prompt on unseen entities. \textbf{(Right)} Knowledge conflict resolution, where the model is queried about trained entities while the context provides conflicting information, and responses reveal the preference between parametric knowledge and in-context knowledge. These behaviors collectively track the co-emergence of dual knowledge capabilities.}
\label{fig:task}
\end{figure*}

In this work, we present the first systematic identification of the training characteristics that enable models to robustly integrate parametric and in-context knowledge. We do so by training models on synthetic corpora~\citep{allenzhu2024physicslanguagemodels31, allenzhu2024physicslanguagemodels32, zucchet2025language} with carefully varied properties to examine how these behaviors emerge and evolve. Specifically, we periodically evaluate three fundamental knowledge-related capabilities throughout the training process: parametric knowledge utilization, in-context knowledge utilization, and knowledge conflict resolution (Figure~\ref{fig:task}).

Our experiments reveal a counterintuitive finding: the robust use of both knowledge sources emerges only when three factors typically regarded as detrimental co-occur. First, \textbf{intra-document repetition} creates the necessary training signal for the co-emergence of parametric and in-context knowledge capabilities (Section~\ref{sec:emerge}). Second, a moderate level of \textbf{intra-document inconsistency} prevents an over-reliance on in-context knowledge (Section~\ref{sec:noise}). Third, a \textbf{skewed knowledge distribution} maintains a balanced reliance on both sources by ensuring that rare facts continue to require in-context knowledge, thereby preventing over-reliance on parametric knowledge (Section~\ref{sec:zipf-impact}). These findings offer empirical guidance for designing data curation for large language models: aggressive preprocessing, such as extensive deduplication and data balancing, may inadvertently impair a model's ability to integrate diverse knowledge sources and resolve conflicts between them.

Our contributions are as follows:
\begin{itemize}[leftmargin=*, itemsep=2pt]
    \item We present the first controlled study examining how specific training data characteristics shape the use of both parametric and in-context knowledge in language models.
    \item We identify that robust use of both knowledge sources emerges when three key factors co-occur: intra-document repetition, intra-document inconsistency, and skewed knowledge distributions.
    \item We demonstrate that these training dynamics generalize to real-world language models (Section~\ref{sec:pythia}) and establish how post-training procedures, such as instruction tuning, further reshape a model's internal arbitration strategies (Section~\ref{sec:posttraining}).
\end{itemize}

\section{Dataset and Setup}
\label{sec:setup}

To study how the three knowledge-related capabilities of language models emerge during training, we design a controlled experimental framework that measures each capability and enables systematic manipulation of training-data properties.
\subsection{Synthetic Biography Dataset}
\label{sec:dataset}

\begin{figure*}[t]
  \centering
  \includegraphics[width=\textwidth]{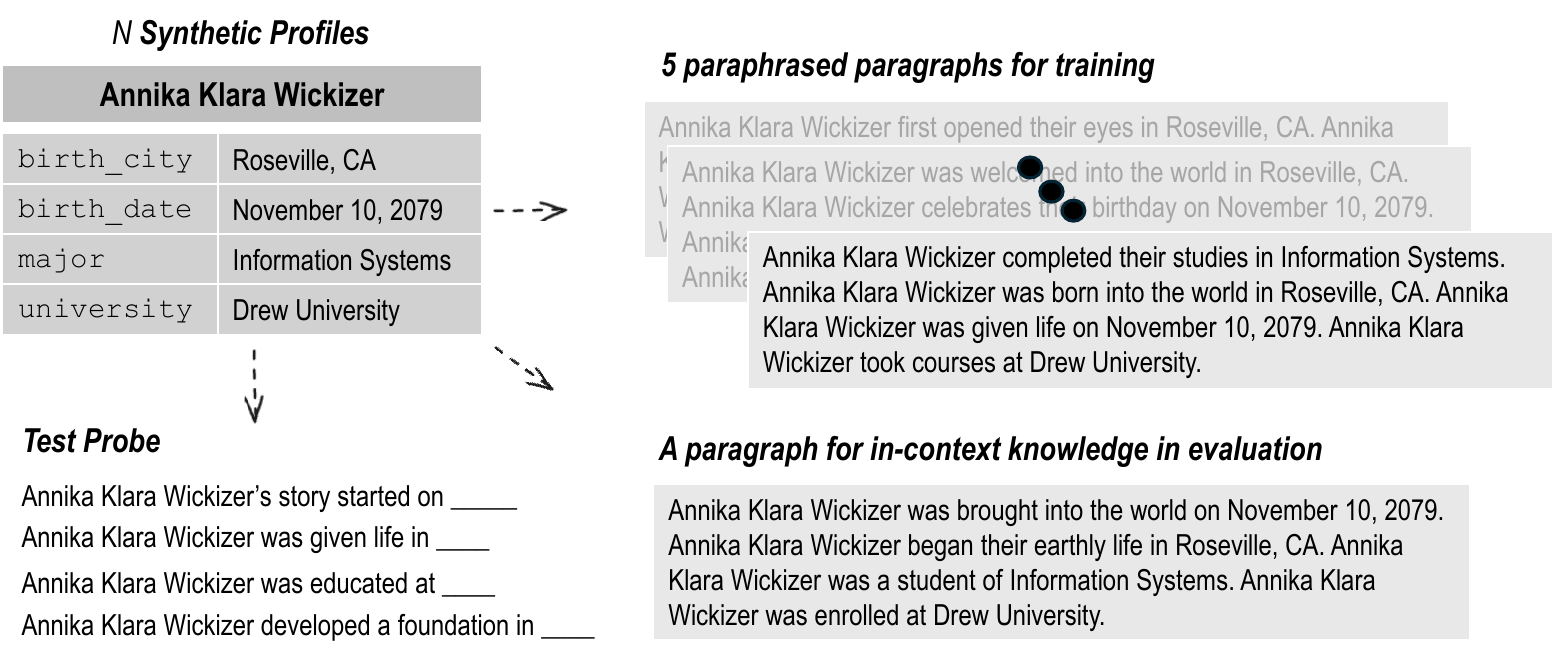}
  \caption{An example from the synthetic biography dataset. Each profile consists of four attributes (\texttt{birth\_date}, \texttt{birth\_city}, \texttt{university}, \texttt{major}), with paragraphs for training, a paragraph for in-context knowledge in evaluation, and test probes for eliciting the model to generate the attributes of each entity.}
  \label{fig:dataset}
\end{figure*}

Following established protocols~~\citep{allenzhu2024physicslanguagemodels31, zucchet2025language}, we construct a synthetic biography dataset in three steps (Figure~\ref{fig:dataset}). First, we generate synthetic entities, each defined by a profile of four attributes: \texttt{birth\_date}, \texttt{birth\_city}, \texttt{university}, and \texttt{major}. Second, for each attribute, we sample 7 distinct surface templates from a finite pool. Third, we instantiate these templates with the attribute values to create biography paragraphs. For each entity, 6 templates per attribute are used to create six paraphrased biography paragraphs: five are reserved as training paragraphs and one is held out as the evaluation context paragraph. The remaining template for each attribute is held out as a cloze-style test probe to obtain the corresponding attribute value. This separation ensures that training paragraphs, evaluation contexts, and test probes never share identical surface forms, encouraging the model to use parametric or in-context knowledge rather than simple string memorization. Detailed specifications are provided in Appendix~\ref{sec:appendix_dataset}.

\subsection{Training Setup}
\label{sec:training_setup}

We train an 8-layer decoder-only Transformer~\citep{vaswani2017attention} from scratch, adopting hyperparameters from prior work~\citep{zucchet2025language} (see Appendix~\ref{app:hyperparams} for details). We use $|\mathcal{E}_{\text{train}}| = 50\text{k}$ entities for training and hold out a separate set of $|\mathcal{E}_{\text{unseen}}| = 50\text{k}$ entities for evaluating in-context knowledge utilization on unseen entities. The model is trained on a corpus of training paragraphs from $\mathcal{E}_{\text{train}}$ using the next-token prediction objective~\citep{radford2018improving}.


\subsection{Evaluation Protocol}
\label{sec:eval_tasks}

We periodically evaluate models in three capabilities described below (Figure~\ref{fig:task}), using exact-match accuracy on 200 randomly sampled entities every 100 training steps.

\paragraph{Parametric Knowledge Utilization (PKU).} 
This metric measures the model's ability to recall learned facts without contextual support. Given an entity $e \in \mathcal{E}_{\text{train}}$ and a test probe $p_a$ for the attribute $a$, the model must generate the correct value $v_a$ solely from its parameters.
\[
\resizebox{0.89\columnwidth}{!}{$
\mathrm{Acc}_{\mathrm{PKU}} 
= \mathop{\mathbb{E}}_{e \sim \mathcal{E}_{\text{train}}} 
\left[ \frac{1}{|A_{e}|} 
\sum_{a \in A_{e}} \mathbf{1}\{M(p_a) = v_a\} \right]
$}
\]
where $A_e$ is the set of attributes for entity $e$, $M(\cdot)$ denotes the model output, and $\mathbf{1}\{\cdot\}$ is the indicator function.

\paragraph{In-Context Knowledge Utilization (ICKU).} 
This metric evaluates the model's ability to extract and utilize knowledge provided in context for entities never seen during training. For $e \in \mathcal{E}_{\text{unseen}}$, we construct a context $C$ containing $e$'s held-out evaluation paragraph along with paragraphs from two other unseen entities as distractors:
\[
\resizebox{0.89\columnwidth}{!}{$
\mathrm{Acc}_{\mathrm{ICKU}} 
= \mathop{\mathbb{E}}_{e \sim \mathcal{E}_{\text{unseen}}} 
\left[ \frac{1}{|A_{e}|} 
\sum_{a \in A_{e}} \mathbf{1}\{M(C, p_a) = v_a\} \right]
$}
\]

\paragraph{Knowledge Conflict Resolution.}
This metric reveals the model's preference when parametric and in-context knowledge conflict. For $e \in \mathcal{E}_{\text{train}}$, we construct a perturbed context $C'_e$ by replacing attribute values with randomly sampled alternatives, then measure how often the model follows each source:
\[
\resizebox{0.89\columnwidth}{!}{$
\mathrm{Pref}_{\mathrm{PK}} = \mathop{\mathbb{E}}_{e \sim \mathcal{E}_{\text{train}}}\left[ \frac{1}{|A_e|}  \sum_{a \in A_e} \mathbf{1}\{M(C'_e, p_a) = v_a\} \right]
$}
\]
\[
\resizebox{0.89\columnwidth}{!}{$
\mathrm{Pref}_{\mathrm{ICK}} = \mathop{\mathbb{E}}_{e \sim \mathcal{E}_{\text{train}}}\left[ \frac{1}{|A_e|} \sum_{a \in A_e} \mathbf{1}\{M(C'_e, p_a) = v'_a\} \right]
$}
\]
where $v_a$ denotes the original (parametric) value and $v'_a$ denotes the perturbed (in-context) value. Higher $\mathrm{Pref}_{\mathrm{PK}}$ indicates stronger reliance on parametric knowledge; higher $\mathrm{Pref}_{\mathrm{ICK}}$ indicates stronger reliance on in-context knowledge. Note that $\mathrm{Pref}_{\mathrm{PK}} + \mathrm{Pref}_{\mathrm{ICK}}$ need not sum to $1$, since the model may also produce outputs matching neither $v_a$ nor $v'_a$.

\section{Experiments}
\label{sec:experiments}

Building on the framework introduced in Section~\ref{sec:setup}, we reverse-engineer how language models acquire these three capabilities (parametric knowledge utilization, in-context knowledge utilization, and confidence-dependent arbitration under knowledge conflict) by systematically manipulating training-data properties and identifying which conditions support the emergence of each capability.

\subsection{Effect of Intra-Document Repetition}
\label{sec:emerge}

\begin{figure*}[t]
\vspace{-8pt}
    \centering
    \includegraphics[width=\textwidth]{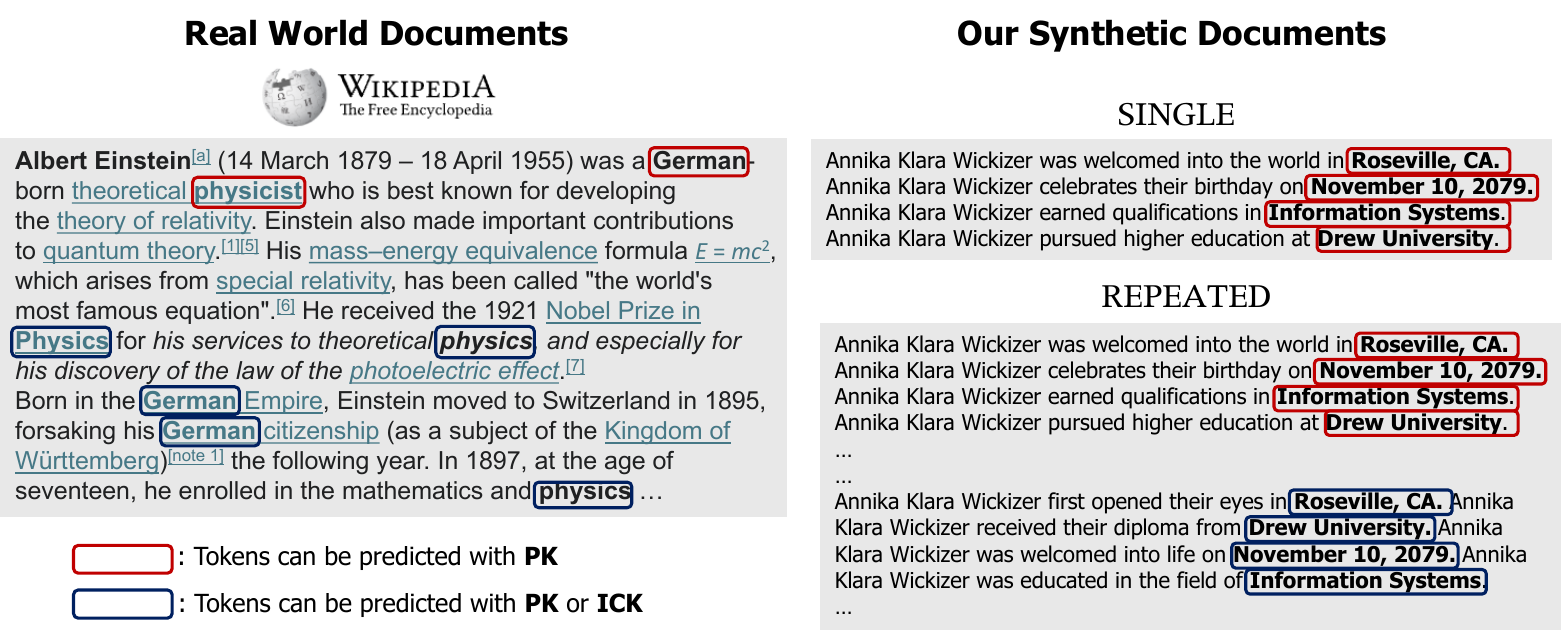}
    \caption{Intra-document repetition in the training corpus. (\textbf{Left}) Real-world Wikipedia text. (\textbf{Right}) Our synthetic corpus variants: \textsc{Single} contains one paragraph per entity, while \textsc{Repeated} contains two paraphrased paragraphs per entity, enabling in-context knowledge utilization on later mentions.}
\label{fig:corpus}
\end{figure*}

\paragraph{Motivation and hypothesis.}
We first examine which factors enable models to use both parametric and in-context knowledge. We hypothesize that intra-document repetition, a common property of natural text in which some information is restated within the same document (Figure~\ref{fig:corpus}), plays a critical role. During the prediction of the next-token, the first mention of a fact requires parametric recall~\citep{geva2023dissecting, meng2022locating}, while later mentions allow the model to take advantage of an earlier context. We hypothesize that this learning signal naturally enables both parametric recall and the use of in-context knowledge.

\paragraph{Design.}
To test this hypothesis, we construct two corpus variants that differ in whether attributes repeat within documents:
\begin{itemize}[leftmargin=*, itemsep=1pt]
    \item \textsc{Single}: Each document contains one paragraph per entity, so attributes appear only once.
    \item \textsc{Repeated}: Each document contains two paraphrased paragraphs per entity. The first mention necessarily relies on parametric knowledge, while the second mention provides an opportunity for the model to use either parametric knowledge or in-context knowledge. To avoid trivial copying based solely on previously mentioned attribute types regardless of the subject, we mix multiple entities within each document. Specifically, we sample two paraphrased paragraphs for each of three distinct entities and shuffle all six paragraphs to form a single training document.
\end{itemize}
In both variants, paragraphs are sampled from the five training paragraphs reserved for each entity in Section~\ref{sec:dataset}.

\begin{figure}[t]
    \centering
    \includegraphics[width=\columnwidth]{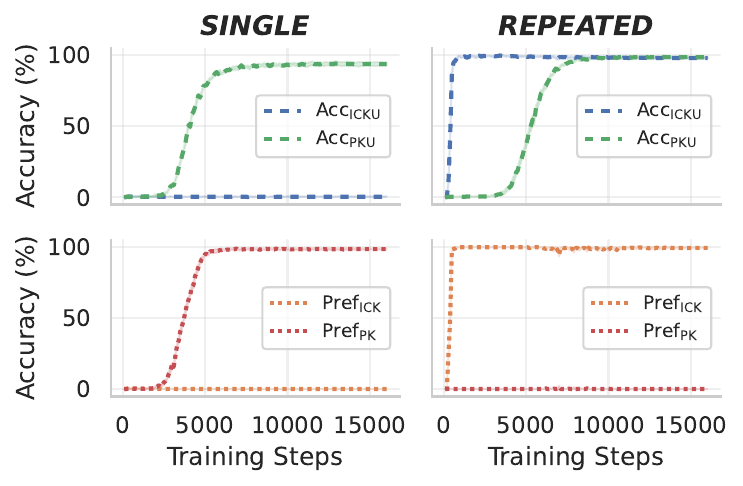}
    \caption{Evaluation results on \textsc{Single} versus \textsc{Repeated}. \textsc{Single} develops only parametric knowledge utilization and always prefers parametric knowledge under conflict. \textsc{Repeated} yields both capabilities, with in-context knowledge utilization emerging first, but consistently prefers in-context knowledge under conflict.}
    \label{fig:result1}
\end{figure}

\paragraph{Finding 1: Repetition yields co-emergence, with in-context knowledge utilization emerging first.}
Figure~\ref{fig:result1} shows that models trained on \textsc{Single} develop only parametric recall, whereas models trained on \textsc{Repeated} acquire both capabilities. Moreover, the ability to use in-context knowledge emerges earlier than parametric recall. One possible explanation for this ordering is a structural asymmetry: in-context knowledge can be used via general copying mechanisms~\citep{olsson2022context}, whereas parametric recall requires jointly learning entity-specific knowledge and a recall mechanism, a combination that develops more gradually, as observed in prior work~\citep{zucchet2025language}.

\paragraph{Finding 2: Clean \textsc{Repeated} corpus induces over-reliance on context.}
Models trained on \textsc{Single} cannot utilize in-context knowledge and therefore trivially prefer parametric knowledge under knowledge conflict. In contrast, models trained on \textsc{Repeated}, despite possessing both capabilities, consistently prefer in-context knowledge under knowledge conflict (Figure~\ref{fig:result1}), even when their parametric knowledge is highly confident. This is evidenced by significantly lower entropy and higher target probability for training entities (see Appendix~\ref{app:confidence}). Such over-reliance on context deviates from the behavior of real-world language models, which tend to prefer parametric knowledge for high-confidence facts~\citep{yu-etal-2023-characterizing, wu2024clasheval}.

\subsection{Effect of Intra-Document Inconsistency}
\label{sec:noise}

\paragraph{Motivation and hypothesis.}
The previous section showed that models trained on clean \textsc{Repeated} data over-rely on in-context knowledge, even when their parametric knowledge is highly confident. This observation raises the question of which properties of natural web corpora discourage such unconditional reliance on in-context knowledge.

We hypothesize that a moderate degree of intra-document inconsistency plays this role. Real-world corpora inevitably contain noise (e.g., typos, imperfect statements, or synonymous paraphrases), making in-context evidence an imperfect signal. When contextual information is occasionally incorrect, the model may learn that parametric knowledge is more reliable for high-confidence facts. To test this hypothesis, we introduce controlled intra-document inconsistency into the training corpus.

\paragraph{Design.}
Starting from \textsc{Repeated}, we inject inconsistency by perturbing the values of entity attributes in the leading paragraph of each document with probability $p \in \{1\%, 5\%, 10\%\}$, replacing them with randomly sampled alternative values, while leaving the later paragraph unchanged (Figure~\ref{fig:noise_figure}).

\begin{figure}[t]
\centering
\includegraphics[width=\columnwidth]{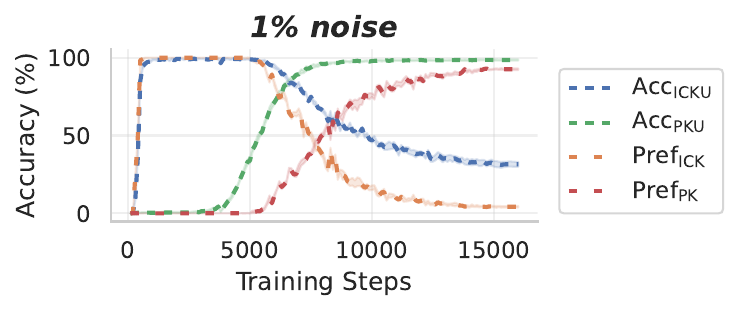}
\caption{Effects of intra-document inconsistency. A 1\% inconsistency rate shifts model preference from in-context to parametric knowledge as training progresses.}
\label{fig:result_noise01}
\end{figure}

\paragraph{Finding 1: Intra-document inconsistency induces a preference transition.}
Figure~\ref{fig:result_noise01} shows a consistent two-stage pattern. Early in training, the ability to use in-context knowledge emerges first, and the model prefers in-context knowledge under conflict. As parametric recall stabilizes, however, the model’s preference gradually shifts toward parametric knowledge. Remarkably, even 1\% inconsistency is sufficient to induce this transition. This behavior suggests that inconsistency imposes an effective ceiling on the reliability of context-based copying; once parametric accuracy exceeds this ceiling, the model increasingly favors parametric knowledge under conflict.

\paragraph{Finding 2: Inconsistency degrades in-context knowledge utilization.}
As the model increasingly relies on parametric knowledge, its ability to use in-context knowledge degrades at convergence. Notably, this degradation occurs even though evaluation is conducted on entities never seen during training: the model increasingly fails to use the provided context when answering questions about entirely unseen entities. One plausible explanation is the gradual forgetting of in-context circuits due to reduced usage~\citep{olsson2022context}. As parametric knowledge becomes more advantageous for frequently observed entities during training, the circuits supporting context use receive diminishing learning signal and gradually deteriorate. Attention analysis in Appendix~\ref{sec:appendix_attention} supports this interpretation: when evaluating on unseen entities, attention initially concentrates on context tokens but progressively shifts toward subject name tokens, suggesting that the circuits responsible for in-context knowledge retrieval are used less during training and gradually forgotten.


\subsection{Effect of Skewed Knowledge Distribution}
\label{sec:zipf-impact}

\paragraph{Motivation and hypothesis.}
To prevent the degradation of in-context knowledge utilization observed in the previous section, we hypothesize that the model must be continuously exposed to predictions that cannot be resolved by parametric knowledge alone during training. In natural web corpora, a vast amount of information exists where some knowledge appears very frequently while most knowledge appears only occasionally (long-tailed knowledge)~\citep{mallen-etal-2023-trust}. We hypothesize that this skewed distribution of knowledge is key to maintaining balanced use of both in-context and parametric knowledge.

\paragraph{Design.}
We construct \textsc{Repeated} corpora where entity occurrences follow a Zipfian distribution with parameter $\alpha = 1$,\footnote{The Zipfian distribution is defined as $P(r) = r^{-\alpha} / \sum_{k=1}^{N} k^{-\alpha}$, where $r$ denotes the frequency rank.} and inject $p=1\%$ inconsistency noise as in the previous section.

\begin{table}[t]
\centering
\small
\begin{tabular}{lcc}
\toprule
Noise & \textbf{Uniform} & \textbf{Zipfian} \\
\midrule
1\%  & 31.5\% & 84.0\% \tiny{(+52.5\%)} \\
5\%  & 16.8\% & 63.9\% \tiny{(+47.1\%)} \\
10\% & 14.1\% & 57.4\% \tiny{(+43.3\%)} \\
\bottomrule
\end{tabular}
\caption{In-context knowledge utilization accuracy at the end of training. \textbf{Uniform} corresponds to the standard \textsc{Repeated} corpus with uniform entity frequency, while \textbf{Zipfian} uses a skewed ($\alpha=1$) entity frequency distribution. Zipfian sampling substantially mitigates degradation relative to uniform sampling under matched inconsistency levels.}
\label{tab:acc_icku_zipf}
\end{table}

\begin{figure}[t]
\centering
\includegraphics[width=\columnwidth]{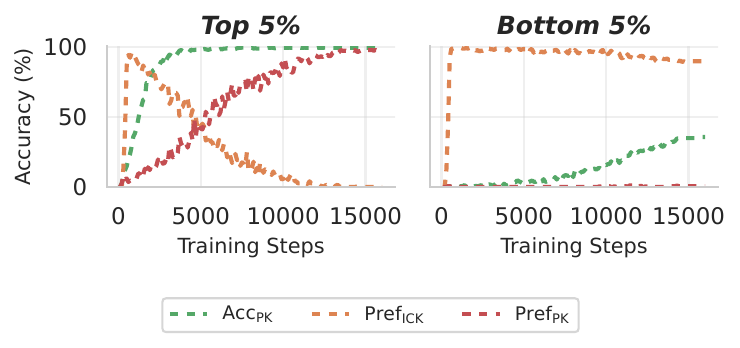}
\caption{Arbitration behavior stratified by entity frequency. High-frequency entities transition to parametric preference, whereas low-frequency entities maintain in-context preference throughout training.}
\label{fig:zipf_arbitration}
\end{figure}

\paragraph{Finding 1: Long-tailed knowledge preserves in-context knowledge utilization capabilities.}
We hypothesized that long-tailed knowledge, for which sufficient parametric knowledge has not accumulated, would require continuous use of in-context knowledge, thereby preventing the degradation of in-context circuits observed in Section~\ref{sec:noise}. Indeed, Table~\ref{tab:acc_icku_zipf} shows substantially less degradation of in-context knowledge utilization under Zipfian distribution across all noise levels. However, when inconsistency noise is too high ($p>1\%$), in-context knowledge utilization appears to degrade too severely; even with Zipfian sampling, the recovery is only partial and final accuracy remains relatively low.

\paragraph{Finding 2: Frequency-dependent arbitration emerges.}
Unlike the uniform setting, where the model's confidence in parametric knowledge saturates uniformly across all trained entities, Zipfian sampling yields varying confidence in parametric knowledge that correlates with entity frequency (Appendix~\ref{app:confidence}, Figure~\ref{fig:zipf_prob}), as observed in real-world language models~\citep{mallen-etal-2023-trust}. This variation enables us to examine how the model's arbitration behavior under knowledge conflicts varies between entities that appear frequently versus infrequently in the training corpus. As shown in Figure~\ref{fig:zipf_arbitration}, high-frequency entities (top 5\% of all entities) transition toward parametric preference as training progresses, as observed in Section~\ref{sec:noise}, while low-frequency entities (bottom 5\% of all entities) maintain in-context preference throughout. Notably, for rare entities we observe that $\mathrm{Acc}_{\mathrm{PKU}}$ exceeds $\mathrm{Pref}_{\mathrm{PK}}$: the model can sometimes answer correctly via parametric recall, yet it still prefers contextual evidence under explicit conflict.

\begin{figure}[t]
\centering
\includegraphics[width=\columnwidth]{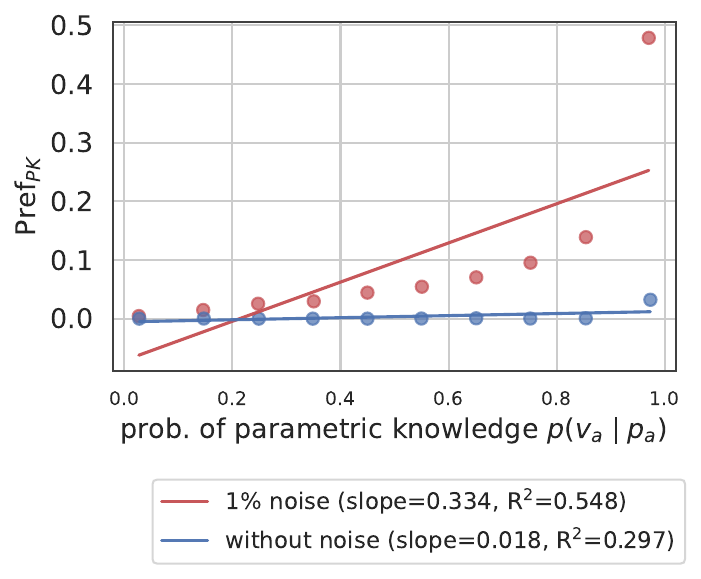}
\caption{Parametric confidence (grouped into 10 bins) against parametric preference under conflict. (\textbf{Red}) With 1\% inconsistency noise, higher confidence yields stronger parametric preference. (\textbf{Blue}) Without inconsistency noise, models show over-reliance on in-context knowledge across all confidence levels.}
\label{fig:zipf_corr}
\end{figure}

\paragraph{Finding 3: Skewed knowledge distribution alone is insufficient.}
So far, we have examined training on data with skewed knowledge distribution in the presence of inconsistency noise. However, does a long-tailed distribution alone, without inconsistency, produce confidence-calibrated arbitration? Figure~\ref{fig:zipf_corr} shows that the answer is no. We measure parametric confidence (grouped into 10 bins) against parametric preference under conflict. We first group the probabilities of predictions on parametric knowledge probes into ten equidistant bins and plot the average $\mathrm{Pref}_{\mathrm{PK}}$ for instances in each bin. Without inconsistency noise, models show a low parametric knowledge preference overall. Only the combination of skewed knowledge distribution and modest inconsistency yields the desired alignment between confidence and preference, where higher confidence leads to stronger preference for parametric knowledge.


\subsection{Summary}
\label{sec:intuition}

Our experiments show that robust use of both knowledge sources emerges when the following three properties of the training corpus co-occur. First, intra-document repetition creates a training signal for both context use and parametric recall, enabling their co-emergence, with the former emerging earlier and the latter following later.

Second, intra-document inconsistency teaches the model not to blindly copy contextual information. Once parametric knowledge has been sufficiently learned and the model is confident in it, the model relies on its parametric prediction rather than defer to context. However, this shift introduces a new problem: the ability to use in-context knowledge degrades as parametric knowledge becomes sufficient for most predictions.

Third, a skewed knowledge distribution resolves this tension through the complementary roles of rare and frequent entities. Rare entities continue to require in-context knowledge, thereby preserving the model's ability to use context throughout training. For frequent entities, whose parametric knowledge is well-learned, the model instead becomes robust to noisy or misleading in-context evidence, producing confidence-dependent arbitration.

When all three properties co-occur, as they naturally do in real web corpora, models develop balanced reliance on both knowledge sources and confidence-dependent arbitration in our controlled setting. Additional hyperparameter experiments, including the number of entities in training data, noise levels, and degree of skewness, are provided in Appendix~\ref{app:additional}.

\section{Discussion and Implications}

\subsection{Do These Dynamics Emerge in Real-World Pretraining?}
\label{sec:pythia}

\begin{figure*}[t]
\centering
\includegraphics[width=\textwidth]{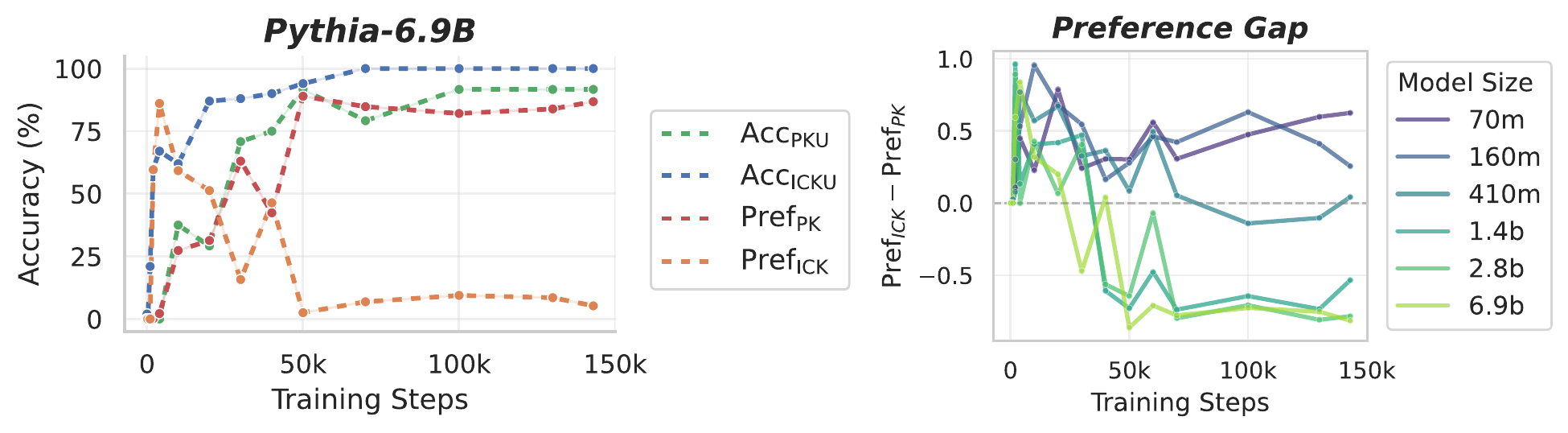}
\caption{Evaluation results of knowledge utilization and conflict resolution in Pythia. (\textbf{Left}) $\mathrm{Acc}_{\mathrm{ICKU}}$, $\mathrm{Acc}_{\mathrm{PKU}}$, $\mathrm{Pref}_{\mathrm{ICK}}$, and $\mathrm{Pref}_{\mathrm{PK}}$ across training steps for Pythia-6.9B. (\textbf{Right}) Preference gap ($\mathrm{Pref}_{\mathrm{ICK}} - \mathrm{Pref}_{\mathrm{PK}}$) across different model sizes, showing a consistent pattern of initial increase followed by decline as training progresses.}
\label{fig:pythia}
\end{figure*}

Our experiments provide a mechanistic account of knowledge utilization, revealing training dynamics that extend beyond previously studied end-state behaviors~\citep{wu2024clasheval, yu-etal-2023-characterizing}. We assess whether similar dynamics arise in real-world pretraining by examining open-source LLMs with publicly available intermediate checkpoints.

We evaluate Pythia~\citep{biderman2023pythia} on parametric knowledge utilization, in-context knowledge utilization, and knowledge conflict resolution at intermediate checkpoints throughout pretraining (experiment details in Appendix~\ref{sec:appendix_pythia}).\footnote{We use Pythia because it provides fine-grained intermediate checkpoints throughout pretraining. Results on another model with publicly available pretraining checkpoints, Olmo~\citep{groeneveld2024olmo}, are provided in Appendix~\ref{sec:appendix_pythia}.} As shown in Figure~\ref{fig:pythia} (left), Pythia exhibits training dynamics consistent with our controlled experiments: the ability to use in-context knowledge emerges earlier than parametric recall, and the model initially prefers in-context knowledge under conflict but gradually shifts toward parametric knowledge, while maintaining high $\mathrm{Acc}_{\mathrm{ICKU}}$ for novel entities throughout training. These results show that the dynamics identified in our controlled experiments also arise during real-world pretraining, providing evidence that our findings capture behavior relevant beyond the synthetic setting.

To examine how this transition scales with model size, we analyze the preference gap ($\mathrm{Pref}_{\mathrm{ICK}} - \mathrm{Pref}_{\mathrm{PK}}$) for models ranging from 70M to 6.9B parameters (Figure~\ref{fig:pythia}, right). All models exhibit a consistent pattern: an initial dominance of in-context knowledge preference gradually shifts toward parametric knowledge preference over training. Notably, larger models show stronger parametric knowledge preference by the end of training, with the preference gap approaching $-1$ for the largest models, consistent with prior observations that larger models tend to rely more heavily on their parametric knowledge~\citep{yu-etal-2023-characterizing}. This trend can be attributed to larger models developing parametric knowledge more effectively, leading to higher confidence in their internal knowledge and consequently a stronger preference for it when conflicts arise.

\subsection{Can Arbitration Strategies Be Reshaped via Post-Training?}
\label{sec:posttraining}

\begin{table}[t]
\small
\centering
\vspace{-5pt}
\begin{tabular}{lcccc}
\toprule
& \multicolumn{2}{c}{Pythia-6.9B} & \multicolumn{2}{c}{Olmo-7B} \\
\cmidrule(lr){2-3} \cmidrule(lr){4-5}
& Pref$_{\text{PK}}$ & Pref$_{\text{ICK}}$ & Pref$_{\text{PK}}$ & Pref$_{\text{ICK}}$ \\
\midrule
Before IT & 0.8677 & 0.0525 & 0.5507 & 0.3894 \\
After IT & 0.1829 & 0.7771 & 0.2137 & 0.7155 \\
\bottomrule
\end{tabular}
\caption{Conflict resolution before and after instruction tuning (IT) for Pythia-6.9B and Olmo-7B. Both models show a shift from parametric knowledge preference to in-context knowledge preference after IT.}
\label{tab:sft}
\vspace{-10pt}
\end{table}

\begin{figure*}[t]
\centering
\includegraphics[width=\textwidth]{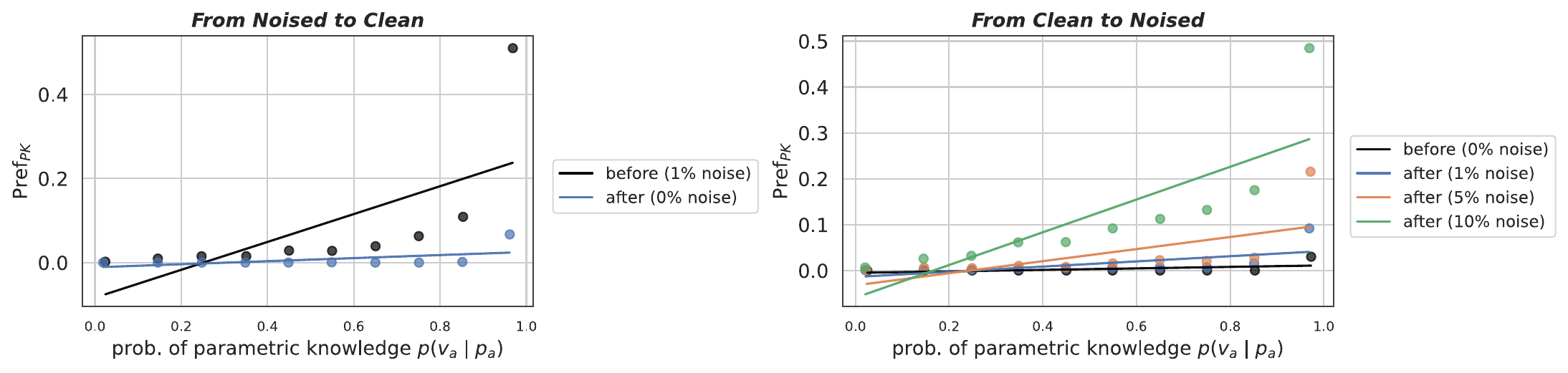}
\caption{Alignment between parametric knowledge confidence and preference under conflict before and after post-training. (\textbf{Left}) Scenario 1: A model pretrained with 1\% noise shows declined parametric knowledge preference after post-training with clean data that has no inconsistency noise. (\textbf{Right}) Scenario 2: A model pretrained without noise initially shows almost no parametric knowledge preference, but after post-training on data with noise, parametric knowledge preference gradually increases according to the model's confidence, with higher noise levels producing stronger confidence-calibrated parametric knowledge preference.}
\label{fig:posttraining}
\end{figure*}

Our findings reveal how the characteristics of the training corpus shape knowledge arbitration strategies during pretraining. A natural question arises: can these strategies be modified after pretraining through post-training procedures such as instruction tuning?

We examine whether instruction tuning affects arbitration behavior in real-world models. Specifically, we evaluate two models at their final pretraining checkpoints (Pythia-6.9B and Olmo-7B) using the same evaluation protocol as in Section~\ref{sec:pythia}, and compare them against their counterparts post-trained on the Tulu dataset~\citep{wang2023tulu}, an instruction-following dataset.

As shown in Table~\ref{tab:sft}, both base models exhibit a higher preference for parametric knowledge. After post-training, however, both models show a reversal: parametric knowledge preference drops while in-context knowledge preference increases. This suggests that instruction tuning, which typically involves data designed to encourage faithful adherence to context, can significantly alter the arbitration strategies established during pretraining.

Having observed that post-training can modify model behavior, we further investigate whether adjusting inconsistency noise in post-training data can by itself control the model's arbitration strategies as intended, given that this noise is a key factor for in-context knowledge reliance in our findings. We conduct post-training on our synthetic Zipfian corpus using answer-only loss with 1,000 entities for 500 steps, substantially smaller in both entity count and training steps than pretraining.

We examine two scenarios: (1) whether a model pretrained with 1\% noise and post-trained on clean data increases its in-context reliance, and (2) whether a model pretrained without noise and post-trained with varying noise levels ($p \in \{1\%, 5\%, 10\%\}$) decreases its in-context reliance.

The results are shown in Figure~\ref{fig:posttraining}. We plot the confidence-preference alignment by binning entities based on parametric knowledge probability and measuring $\mathrm{Pref}_{\mathrm{PK}}$ for each bin. The results show that adjusting noise levels alone can reshape this alignment. More broadly, noise level is one instance of a more general principle: controlling how much the training data allows the model to rely on context. This calibration of context reliability directly shapes the model's arbitration strategy between parametric and in-context knowledge. In practice, it offers a concrete lever for deployment: raising the context noise in post-training data encourages parametric reliance when retrieval sources are unreliable (e.g., web-based search), while lowering it encourages in-context reliance in domains with highly curated retrieval sources (e.g., legal or medical databases).

\section{Related Work}

\paragraph{Knowledge Utilization and Conflict Resolution in Language Models.}
Large language models store factual knowledge in their parameters during pretraining~\citep{roberts2020much, petroni2019language, geva2020transformer} and can also leverage in-context knowledge at inference time without explicit fine-tuning~\citep{lewis2021retrieval, ram2023incontext, shi2023replug, mallen-etal-2023-trust, ram-etal-2023-context}. When these sources conflict~\citep{neeman2022disentqa}, models exhibit confidence-dependent arbitration, preferring parametric knowledge for well-learned facts while deferring to in-context knowledge for less familiar information~\citep{wu2024clasheval, yu-etal-2023-characterizing, lee2026judgingreferenceuncoveringknowledgedriven}. Several methods have been proposed to steer this behavior through attention manipulation or contrastive decoding~\citep{litaming, yu-etal-2023-characterizing, sun2025redeep, jin-etal-2024-cutting}. However, prior work in this line focuses on analyzing or modifying post-hoc behavior~\citep{kortukov2024studyinglargelanguagemodel, xie2023adaptive, longpre2021entity}, leaving open the question of how these capabilities emerge during training.

\paragraph{Experiments with Controlled Training Data.}
Several recent works use controlled training setups to disentangle how data properties shape what models learn. For instance, \citet{chan2022data, singh2023transient} investigate how data properties enable in-context and in-weight learning to co-exist in transformer-based classifiers. However, their work is limited to classification tasks, which may exhibit different dynamics from language models trained with next-token prediction. Other works present controlled studies with synthetic corpora to investigate parametric knowledge acquisition in language models~\citep{allenzhu2024physicslanguagemodels31, allenzhu2024physicslanguagemodels32, zucchet2025language, kim2026bilinearrepresentationmitigatesreversal}, but do not address in-context utilization~\citep{olsson2022context} or conflict resolution~\citep{wu2024clasheval}. We extend these directions by examining how training-data characteristics shape both the co-emergence of parametric and in-context knowledge utilization and the development of conflict arbitration strategies.

\section{Conclusion}

We presented a systematic analysis of how training data characteristics shape the use of parametric and in-context knowledge in language models through controlled experiments. We identified a counterintuitive finding: robust use of both knowledge sources emerges when three properties commonly regarded as detrimental co-occur, including intra-document repetition, intra-document inconsistency, and skewed knowledge distribution. Experiments on real-world language models show that similar dynamics arise during pretraining, and our post-training experiments demonstrate that knowledge arbitration strategies can be reshaped by adjusting data characteristics. These findings offer practical guidance for designing training data that supports balanced use of parametric and in-context knowledge in language models.

\section{Limitations}

Our study primarily relies on a synthetic biography dataset and small-scale models. Research based on synthetic data offers the critical advantage of isolating and controlling complex factors while enabling numerous reproducible experiments, which would be infeasible in realistic scenarios. This design choice, however, naturally raises questions about the generalizability of our findings. At the same time, independently controlling properties such as repetition, noise, and frequency distribution in real-world training data is highly challenging, and repeatedly pretraining large-scale language models to isolate each factor is prohibitively costly, which limits the feasibility of causal analysis in such settings. Accordingly, we adopt a complementary structure: establishing causal relationships through controlled experiments in the synthetic environment, and then showing that similar patterns arise as correlational evidence in real-world models whose pretraining checkpoints are publicly available. Through this approach, we provide a systematic understanding of which training data properties shape a model's knowledge utilization behavior, and we believe our work can serve as a foundation for follow-up studies in more realistic, larger-scale scenarios that are otherwise difficult to control. We hope our findings will be further extended and validated in such settings in future work.

\section*{Acknowledgments}

This work was supported by the National Research Foundation of Korea (NRF) grant funded by the Korea government (MSIT)(RS-2024-00348233). K. Jung is with ASRI, Seoul National University, Korea. The Institute of Engineering Research at Seoul National University provided computing resources.


\bibliography{custom}

\appendix

\section{Biography Dataset Construction}
\label{sec:appendix_dataset}

Following prior work~\citep{allenzhu2024physicslanguagemodels31, zucchet2025language}, we first construct $N$ synthetic person profiles. Each profile contains four attributes: \texttt{birth\_date}, \texttt{birth\_city}, \texttt{university}, and \texttt{major}. Names (first/middle/last) are sampled by randomly composing entries from a public name database.\footnote{\url{https://github.com/smashew/NameDatabases/tree/master/NamesDatabases}}
For \texttt{birth\_date}, we sample a date uniformly between 1900 and 2099. For \texttt{birth\_city} and \texttt{university}, we sample from curated lists of 200 values each, and for \texttt{major} from a list of 100 values, all derived from Wikipedia.\footnote{\url{https://en.wikipedia.org/wiki/}}
For each attribute, we sample 7 distinct surface templates from a finite template pool. An example of template for \texttt{birth\_date} is shown below.

\begin{tcolorbox}[enhanced, colback=white, colframe=black, title={An example of templates for \texttt{birth\_date}}]
\small
\begin{enumerate}[leftmargin=*, itemsep=0pt, topsep=0pt]
\tiny{
\item \texttt{{person} was born on {birth\_date}.}
\item \texttt{{person} came into the world on {birth\_date}.}
\item \texttt{{person} entered this world on {birth\_date}.}
\item \texttt{{person} was brought into the world on {birth\_date}.}
\item \texttt{{person} took their first breath on {birth\_date}.}
\item \texttt{{person} began their life journey on {birth\_date}.}
\item \texttt{{person} celebrates their birthday on {birth\_date}.}
\item \texttt{{person} first opened their eyes on {birth\_date}.}
\item \texttt{{person} was welcomed into life on {birth\_date}.}
\item \texttt{{person} arrived on {birth\_date}.}
\item \texttt{{person}'s story started on {birth\_date}.}
\item \texttt{{person} was born to the world on {birth\_date}.}
\item \texttt{{person} was delivered into the world on {birth\_date}.}
\item \texttt{{person} was given life on {birth\_date}.}
\item \texttt{{person} was welcomed into the world on {birth\_date}.}
\item \texttt{{person} began their journey on Earth on {birth\_date}.}
\item \texttt{{person} made their debut in the world on {birth\_date}.}
\item \texttt{{person} became a part of the world on {birth\_date}.}
\item \texttt{{person} was born into this life on {birth\_date}.}
\item \texttt{{person} came to life on {birth\_date}.}
}
\end{enumerate}
\end{tcolorbox}

\section{Training Details}
\label{app:hyperparams}

For our controlled experiments, we use a decoder-only Transformer following the GPT-2 architecture\footnote{\url{https://huggingface.co/openai-community/gpt2}}. The model configuration is summarized in Table~\ref{tab:model-config}. The training hyperparameters are listed in Table~\ref{tab:train-hparams}. All experiments are implemented using the \texttt{Hugging Face TRL} library\footnote{\url{https://huggingface.co/docs/trl/index}} and conducted on a single NVIDIA A100 GPU. Each training run requires approximately four to six hours.

\begin{table}[h]
\centering
\begin{tabular}{ll}
\toprule
Component & Value \\
\midrule
Embedding dimension & 512 \\
Layers & 8 \\
Attention heads & 8 \\
FFN inner dimension & 2048 \\
Context length & 512 \\
\bottomrule
\end{tabular}
\caption{Model architecture.}
\label{tab:model-config}
\end{table}

\begin{table}[h]
\centering
\begin{tabular}{ll}
\toprule
Hyperparameter & Value \\
\midrule
Max training steps & 16{,}000 \\
Batch size & 128 \\
Learning rate & $4 \times 10^{-4}$ \\
Weight decay & 0.10 \\
LR scheduler & Cosine \\
Sequence length & 512 \\
Numerical precision & bfloat16 \\
\bottomrule
\end{tabular}
\caption{Training hyperparameters.}
\label{tab:train-hparams}
\end{table}

\section{Intra-Document Inconsistency}

\begin{figure}[h]
    \centering
    \includegraphics[width=\linewidth]{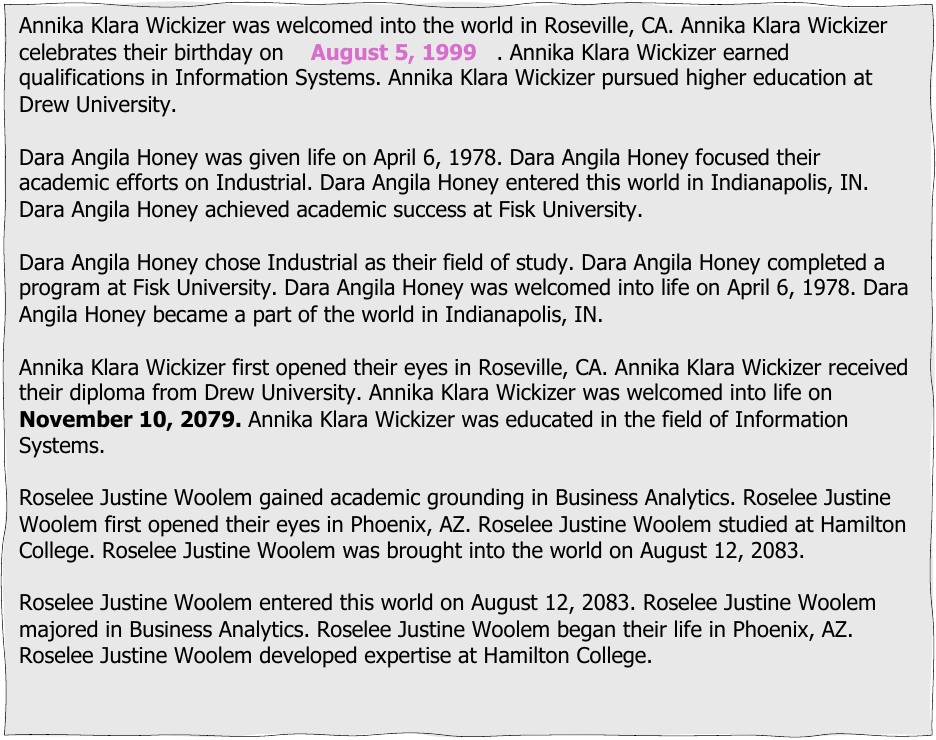}
    \caption{Example of a document with injected inconsistency noise. The value highlighted in pink was injected as noise with some probability and therefore does not match the later unperturbed value, “November 10, 2079.”}
    \label{fig:noise_figure}
\end{figure}

Figure~\ref{fig:noise_figure} illustrates a document from the \textsc{Repeated} corpus in which intra-document inconsistency has been injected.

\section{Confidence for Parametric Knowledge}
\label{app:confidence}

We measure two key metrics at the final token position of each test probe: (1) the probability assigned to the target token, and (2) the entropy of the probability distribution over the vocabulary. These metrics provide complementary perspectives on model confidence: the target probability reflects how strongly the model predicts the correct answer, while the entropy captures the overall uncertainty in the prediction.

\begin{table}[h]
\centering
\vspace{-5pt}
\resizebox{\columnwidth}{!}{%
\begin{tabular}{lcccc}
\toprule
& \multicolumn{2}{c}{$\mathcal{E}_{\text{train}}$} & \multicolumn{2}{c}{$\mathcal{E}_{\text{unseen}}$} \\
\cmidrule(lr){2-3} \cmidrule(lr){4-5}
& 0\% noise & 1\% noise & 0\% noise & 1\% noise \\
\midrule
Target prob. & 0.998 & 0.997 & 0.024 & 0.034 \\
Entropy (nats) & 0.011 & 0.016 & 0.955 & 1.236 \\
\bottomrule
\end{tabular}}%
\caption{Target token probability and entropy measured at the last token of the test probe by entities}
\label{tab:entropy}
\end{table}

Table~\ref{tab:entropy} presents these measurements for entities in both $\mathcal{E}_{\text{train}}$ (seen during training) and $\mathcal{E}_{\text{unseen}}$ (held-out entities) under two training conditions: without noise and with 1\% inconsistency noise. For $\mathcal{E}_{\text{train}}$, the model exhibits extremely high confidence, assigning near-perfect probability to target tokens with very low entropy. This indicates that the model has successfully acquired and can reliably retrieve parametric knowledge for entities it encountered during training. In contrast, for $\mathcal{E}_{\text{unseen}}$, the model shows substantially lower confidence, with lower target probabilities and much higher entropy values.

\begin{figure}[t]
\centering
\includegraphics[width=\linewidth]{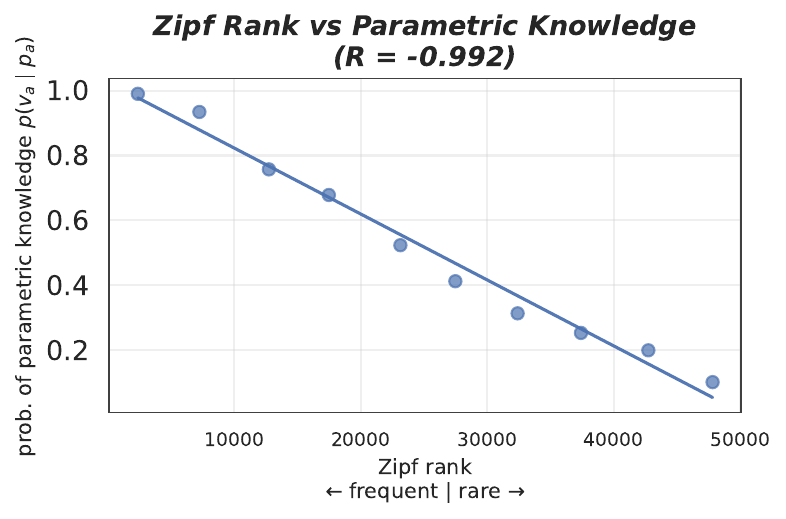}
\caption{Relationship between entity frequency (Zipf rank) and parametric knowledge strength. The model exhibits a strong negative correlation between Zipf rank and the probability $p(v_a \mid p_a)$ of generating the correct attribute value given only the probe prompt. More frequent entities have stronger parametric knowledge while rare entities show weaker parametric knowledge.}
\label{fig:zipf_prob}
\end{figure}

\paragraph{Parametric knowledge varies with entity frequency.}
We investigate how the strength of parametric knowledge varies across entities in $\mathcal{E}_{\text{train}}$ as a function of their frequency in the pretraining corpus. Figure~\ref{fig:zipf_prob} shows the relationship between Zipf rank (where lower ranks indicate more frequent entities) and the probability $p(v_a \mid p_a)$ assigned to the correct attribute value when given only the probe prompt $p_a$, averaged across all attributes. The results reveal a strong negative correlation, demonstrating that parametric knowledge strength is tightly coupled with entity frequency.

\section{Attention Pattern Analysis}
\label{sec:appendix_attention}

We analyze attention patterns to investigate the mechanisms underlying the degradation of in-context knowledge utilization observed in Section~\ref{sec:noise}. By examining the model trained on the \textsc{Repeated} corpus with 1\% inconsistency, we indirectly examine the circuits used for parametric versus in-context knowledge utilization.

\begin{figure}[t]
\centering
\includegraphics[width=\linewidth]{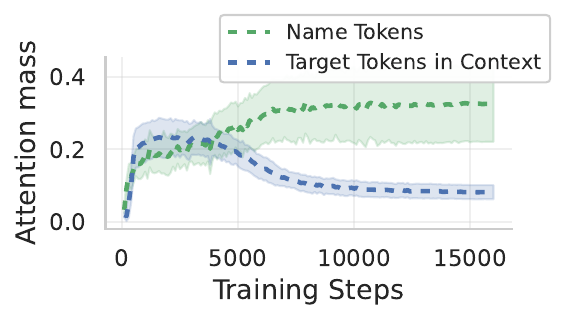}
\caption{Changes in the layer-wise sum of attention mass at the last token of the test probe when the model trained with 1\% noise performs in-context knowledge utilization for $\mathcal{E}_{\text{unseen}}$ entities. \textbf{Green} indicates the attention allocated to name tokens in the test probe, while \textbf{blue} indicates the attention allocated to target tokens in the context.}
\label{fig:result_attention}
\end{figure}

We analyzed the attention patterns at the last token position of the test probe during in-context knowledge utilization for $\mathcal{E}_{\text{unseen}}$ entities. Figure~\ref{fig:result_attention} shows the layer-wise sum of attention mass over the course of training. We distinguish between two types of attention targets: (1) name tokens in the test probe (shown in green), which are more associated with parametric knowledge retrieval~\citep{meng2022locating, zucchet2025language, geva2023dissecting}, and (2) target attribute tokens in the context (shown in blue), which are needed for in-context knowledge utilization~\citep{olsson2022context}.

We find that early in training, attention is heavily concentrated on target attribute tokens in the context, consistent with successful in-context knowledge use mediated by in-context induction circuits~\citep{olsson2022context}. However, as training progresses and parametric knowledge use stabilizes, attention gradually shifts toward name tokens in the test probe. Notably, this shift occurs even when evaluating on $\mathcal{E}_{\text{unseen}}$ entities, for which the model has no parametric knowledge (see Table~\ref{tab:entropy}).

We hypothesize that the presence of inconsistency noise during training introduces imperfection in in-context knowledge utilization, making contextual information a less reliable signal. As a result, once parametric knowledge becomes sufficiently stable, the model increasingly defaults to parametric knowledge retrieval across all situations. Consequently, in-context knowledge utilization circuits receive progressively less training signal. In combination with regularization effects such as weight decay~\citep{loshchilov2017decoupled}, this reduced usage leads to a gradual degradation of the model’s ability to utilize in-context knowledge over the course of training.

This analysis helps explain how a skewed knowledge distribution (Section~\ref{sec:zipf-impact}) can preserve in-context knowledge utilization. The continuous presence of unfamiliar or low-frequency entities in the training distribution forces the model to repeatedly rely on in-context knowledge, thereby preventing the complete abandonment of in-context knowledge circuits.

\section{Evaluation on Real-World LLMs}
\label{sec:appendix_pythia}

\begin{figure*}[h]
\centering
\includegraphics[width=\linewidth]{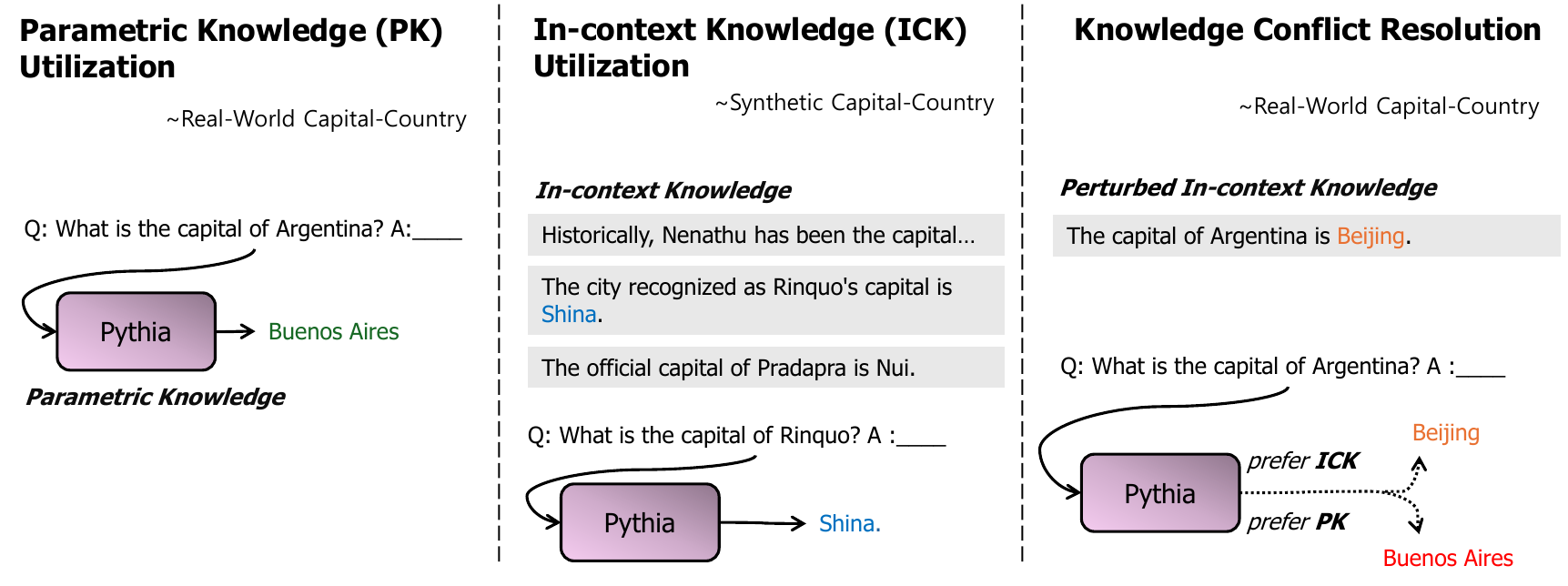}
\caption{Three knowledge utilization scenarios in real-world large language models. (\textbf{Left}) Parametric knowledge utilization, where the model recalls country and capital facts from real-world data that were encoded in its parameters during training. (\textbf{Middle}) In-context knowledge utilization, where the model relies on synthetic country and capital pairs provided only in the context. (\textbf{Right}) Knowledge conflict resolution, where the model is queried about real-world countries while the prompt supplies perturbed (incorrect) capitals, allowing us to examine whether the model prefers parametric knowledge or the perturbed in-context knowledge.}
\label{fig:pythia_task}
\end{figure*}

We adapt the evaluation scenarios used in our controlled experiments to settings applicable to large language models trained on real-world web corpora. Since such corpora contain abundant information about countries and their capitals, we define the set of training entities $\mathcal{E}_{\text{train}}$ as Real-World Countries and evaluate whether the model can correctly predict their corresponding capital cities.

To this end, we construct a Real-World Country and Capital Set based on the country and capital pairs introduced in \citet{hernandez2023linearity}. We then build question and answer style probes as illustrated in Figure~\ref{fig:pythia_task} and define the \textbf{Parametric Knowledge Utilization} (PKU) scenario. We measure $\mathrm{Acc}_\mathrm{PKU}$ by checking whether the correct capital appears within the first 64 generated tokens.

For the \textbf{In-Context Knowledge Utilization} (ICKU) scenario, we evaluate the model's ability to use knowledge provided only in the prompt context. We construct 100 artificial country and capital pairs that do not correspond to any real-world entities, forming a Synthetic Country and Capital Set. These pairs are provided only within the prompt context, and $\mathrm{Acc}_\mathrm{ICKU}$ is computed by verifying whether the correct synthetic capital is generated within 64 tokens.

Finally, for \textbf{Knowledge Conflict Resolution}, we perturb the in-context knowledge by replacing the true capitals in the Real-World Country and Capital Set with incorrect alternatives. Given these perturbed contexts and the corresponding test probes, we evaluate whether the model follows the in-context knowledge or instead relies on its parametric knowledge. This allows us to compute $\mathrm{Pref}_\mathrm{ICK}$ and $\mathrm{Pref}_\mathrm{PK}$, reflecting the model's preference under explicit knowledge conflict.

\begin{figure}[h]
\centering
\includegraphics[width=\linewidth]{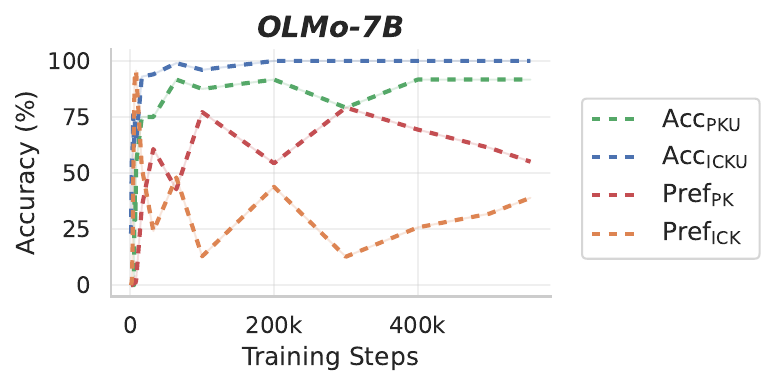}
\caption{Evaluation results of knowledge utilization and conflict resolution in Olmo-7B.}
\label{fig:olmo}
\end{figure}

Extending the results with the Pythia models in Section~\ref{sec:pythia}, we conduct the same set of experiments on Olmo-7B~\citep{groeneveld2024olmo}. As shown in Figure~\ref{fig:olmo}, Olmo-7B exhibits qualitatively similar patterns to those observed in Pythia: in-context knowledge utilization emerges earlier, followed by the stabilization of parametric knowledge utilization, and preference shifts in resolving conflicts between parametric and in-context knowledge. These results suggest that the knowledge utilization dynamics identified in our analysis are not specific to a particular model family, but also appear across different large-scale language models trained on real-world data.

\section{Additional Experimental Results}
\label{app:additional}
We further characterize the influence of training data; unless otherwise noted, all experiments are conducted on the \textsc{Repeated} corpus.

\subsection{Effect of the Number of Training Entities}
\label{app:entity-ablation}

\begin{figure*}[t]
    \centering
    \begin{subfigure}[b]{\textwidth}
        \centering
        \includegraphics[width=\textwidth]{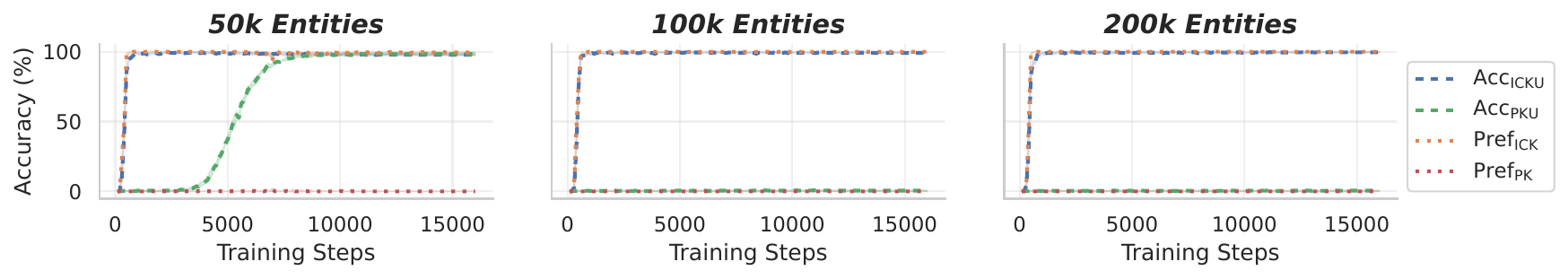}
        \caption{Effect of the number of training entities.}
        \label{fig:entity_ablation}
    \end{subfigure}
    \\[6pt]
    \begin{subfigure}[b]{\textwidth}
        \centering
        \includegraphics[width=\textwidth]{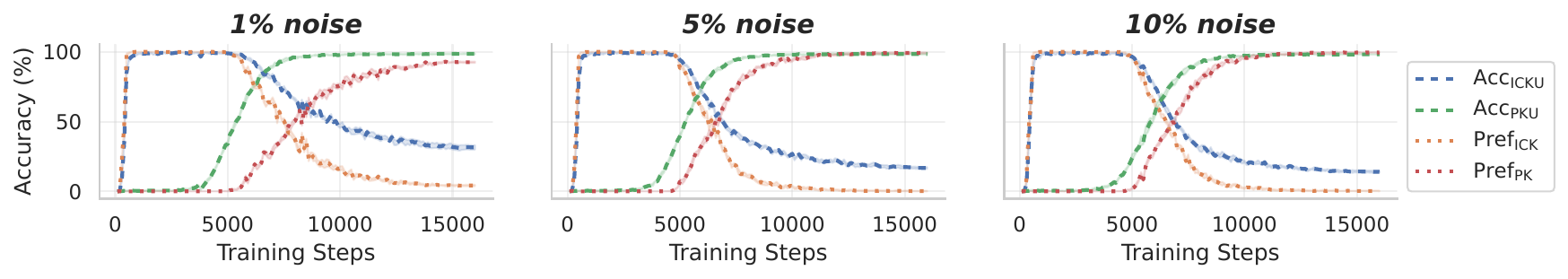}
        \caption{Effect of intra-document inconsistency noise.}
        \label{fig:noise_ablation}
    \end{subfigure}
    \\[6pt]
    \begin{subfigure}[b]{\textwidth}
        \centering
        \includegraphics[width=\textwidth]{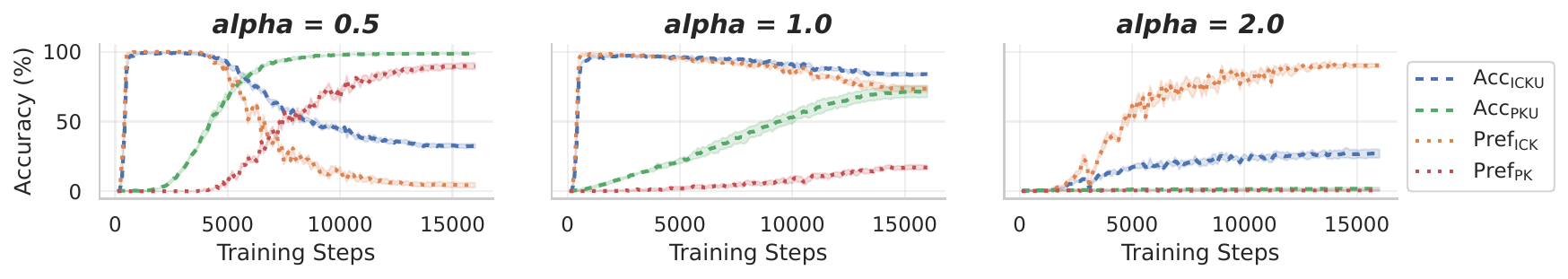}
        \caption{Effect of knowledge distribution (Zipf exponent $\alpha$).}
        \label{fig:skewness_ablation}
    \end{subfigure}
    \caption{Evaluation results during training of in-context knowledge utilization ($\mathrm{Acc}_{\mathrm{ICKU}}$), parametric knowledge utilization ($\mathrm{Acc}_{\mathrm{PKU}}$), and knowledge conflict preferences ($\mathrm{Pref}_{\mathrm{ICK}}$, $\mathrm{Pref}_{\mathrm{PK}}$) under varying corpus properties.}
    \label{fig:ablation_combined}
\end{figure*}

\figurename~\ref{fig:entity_ablation} compares \textsc{Repeated} runs with 50k, 100k, and 200k training entities. With 50k entities, both in-context knowledge utilization ($\mathrm{Acc}_\mathrm{ICKU}$) and parametric knowledge utilization ($\mathrm{Acc}_\mathrm{PKU}$) emerge, with $\mathrm{Acc}_\mathrm{ICKU}$ emerging earlier and $\mathrm{Acc}_\mathrm{PKU}$ following as training stabilizes. In contrast, for 100k and 200k entities, $\mathrm{Acc}_\mathrm{PKU}$ fails to rise: the model learns to use in-context knowledge but does not develop robust parametric utilization. This suggests that when the number of entities exceeds the model's capacity, parametric knowledge cannot be stably acquired, leaving in-context knowledge utilization as the dominant strategy.

\figurename~\ref{fig:noise_ablation} examines training dynamics under intra-document inconsistency levels of $1\%$, $5\%$, and $10\%$. Even $1\%$ noise is sufficient to induce a phase shift in conflict-time preference: as $\mathrm{Acc}_\mathrm{PKU}$ stabilizes, the model transitions from preferring in-context knowledge ($\mathrm{Pref}_\mathrm{ICK}$) to preferring parametric knowledge ($\mathrm{Pref}_\mathrm{PK}$). Increasing noise accelerates this shift but also degrades $\mathrm{Acc}_\mathrm{ICKU}$ at convergence, indicating over-reliance on parametric knowledge and a reduced ability to use in-context knowledge.

\figurename~\ref{fig:skewness_ablation} examines training dynamics under Zipfian sampling with $\alpha \in \{0.5, 1.0, 2.0\}$. A near-uniform regime ($\alpha{=}0.5$) yields progressive degradation of $\mathrm{Acc}_\mathrm{ICKU}$ over training, consistent with the model drifting toward parametric recall even for unfamiliar entities. An overly skewed regime ($\alpha{=}2.0$) produces undesirable dynamics in which parametric utilization fails to activate, because most entities appear too rarely to accumulate sufficient parametric learning signal. A moderate skew ($\alpha{=}1.0$) best preserves $\mathrm{Acc}_\mathrm{ICKU}$ for rare or novel entities while still supporting stable $\mathrm{Acc}_\mathrm{PKU}$ and a robust preference for parametric knowledge on frequently seen facts.

\section{Controlling In-Context Preference via Attention Head Amplification}
\label{app:icl_heads}

We investigate whether the arbitration strategies identified in our controlled experiments can be further modulated at inference time through attention-head manipulation, providing a complementary view to our training-data analysis. Following \citet{yu-etal-2023-characterizing}, we identify in-context heads, namely attention heads that contribute most to in-context knowledge utilization, and scale their activations by a factor $\alpha$ to amplify or diminish their influence during knowledge conflicts.

We apply this intervention to two models trained under different corpus conditions on the \textsc{Repeated} corpus with Zipfian distribution: one trained without noise, and one trained with $1\%$ inconsistency noise. We then measure conflict-time preferences for high-frequency entities across $\alpha \in \{0.0, 0.5, 1.0, 2.0\}$ (Table~\ref{tab:alpha}).

\begin{table}[h]
\centering
\resizebox{\columnwidth}{!}{%
\begin{tabular}{lcccc}
\toprule
& \multicolumn{2}{c}{\textbf{Zipf, 0\% noise}} & \multicolumn{2}{c}{\textbf{Zipf, 1\% noise}} \\
\cmidrule(lr){2-3} \cmidrule(lr){4-5}
$\alpha$ & $\mathrm{Pref}_{\mathrm{PK}}$ (\%) & $\mathrm{Pref}_{\mathrm{ICK}}$ (\%) & $\mathrm{Pref}_{\mathrm{PK}}$ (\%) & $\mathrm{Pref}_{\mathrm{ICK}}$ (\%) \\
\midrule
0.0 & 88.00 & 7.00  & 100.00 & 0.00 \\
0.5 & 42.75 & 40.00 & 98.25  & 0.50 \\
1.0 & 19.50 & 69.50 & 97.00  & 1.75 \\
2.0 & 9.25  & 81.75 & 74.75  & 17.50 \\
\bottomrule
\end{tabular}}%
\caption{Effect of in-context head amplification on conflict-time preference for high-frequency entities. $\alpha$ scales the activations of identified in-context heads.}
\label{tab:alpha}
\end{table}

Both models show that in-context preference can be steered through $\alpha$, consistent with observations on real-world models. However, the response magnitude differs substantially with training-data characteristics: the model trained with $1\%$ noise exhibits a much stronger baseline preference for parametric knowledge and requires large amplification ($\alpha=2$) to noticeably elicit in-context behavior, whereas the noise-free model transitions toward in-context preference even at $\alpha=0.5$. This reflects the training-data-induced arbitration bias identified in our main experiments, in which the noisy pretraining corpus instills a persistent reliance on parametric knowledge that resists simple inference-time intervention. Since attention-head manipulation can also adversely affect the model's general capabilities~\citep{litaming}, our training-data analysis and post-hoc inference-time interventions are best viewed as complementary rather than substitutes.

\section{The Use of Large Language Models}
We used large language models to assist with the preparation of this paper. Specifically, they were employed for writing support, including grammar correction, wording refinement, and minor stylistic edits, as well as for developing code used in the experiments.

\end{document}